\documentclass{article}
\usepackage[utf8]{inputenc}
\usepackage[T1]{fontenc}    
\usepackage[english]{babel}
\usepackage{biblatex, graphicx, booktabs}
\usepackage{arxiv}
\usepackage[section]{placeins}
\title{A Machine Learning Analysis of the Features in Deceptive and Credible News}
\author{
  Qi Jia (Jerry) Sun\thanks{jerrys7.github.io} \\
  Discourse Processing Lab\\
  Simon Fraser University\\
  Burnaby, BC, Canada \\
  \texttt{qjsun@sfu.ca} \\
}

\begin{document}
\maketitle

\begin{abstract}
Fake news is a type of pervasive propaganda that spreads misinformation online, taking advantage of social media’s extensive reach to manipulate public perception. Over the past three years, fake news has become a focal discussion point in the media due to its impact on the 2016 U.S. Presidential election. Fake news can have severe real-world implications: in 2016, a man walked into a pizzeria carrying a rifle because he read that “Hillary Clinton was harboring children as sex slaves”. This project presents a high accuracy (87\%) machine learning classifier that determines the validity of news based on the word distributions and specific linguistic and stylistic differences in the first few sentences of an article. This can help readers identify the validity of an article by looking for specific features in the opening lines aiding them in making informed decisions. Using a dataset of 2,107 articles from 30 different websites, this project establishes an understanding of the variations between fake and credible news by examining the model, dataset, and features. This classifier appears to use the differences in word distribution, levels of tone authenticity, and frequency of adverbs, adjectives, and nouns. The differentiation in the features of these articles can be used to improve future classifiers. This classifier can also be further applied directly to browsers as a Google Chrome extension or as a filter for social media outlets or news websites to reduce the spread of misinformation.
\end{abstract}


\section{Introduction}
\subsection{Fake News}
\large 
Fake news is a type of pervasive propaganda that spreads misinformation online, taking advantage of social media’s extensive reach to manipulate public perception. Over the past three years, fake news has become a focal discussion point in the media due to its impact on the 2016 U.S. Presidential election. Fake news can have severe real-world implications: for instance, in 2016, a man walked into a pizzeria carrying a rifle because he read online that “Hillary Clinton was harboring children as sex slaves”. [9] 
\subsection{Machine Learning}
Machine learning (ML) is a system that utilizes a statistical model consisting of a predetermined set of features and algorithms to make predictions on unseen data without supervision.  ML algorithms have been widely adopted for spam email filtering. Parallels can be drawn between spam emails and fake news such as grammatical errors, misinformation, limited vocabulary, and their purpose of manipulating the reader. Thus, applying a similar ML method from spam emails to fake news could yield promising results. 
\subsection{Purpose}
This project seeks to find a ML classifier that determines the validity of news based on the word distributions and specific linguistic and stylistic differences of the first few sentences of an article. This can help readers identify the validity of an article early on by looking for specific features in the opening lines and differentiate fake from real news. \\ \newline 
Using a dataset of 2,107 articles from 30 different websites, this project seeks to establish an understanding of the variations between fake and credible news by examining the model, dataset, and linguistic features. The differentiation in the features of these articles can be used to improve future classifiers. A deeper understanding of the differences between deceptive and credible media will further the collective progress in the battle against fake news. 

\section{Related Work}
In December 2016, the fake news challenge was launched (www.fakenewschallenge.org). Its purpose was to explore how ML could combat fake news. The challenge was to determine the level of agreement between a statement and its headline. This challenge increased the usage of ML in fake news, and within the past few years, many datasets were prepared for fake news classification such as the L.I.A.R. dataset [20], which categorizes news into 5 levels of credibility, Rashkin et al., 2017 dataset [15], a dataset differentiating between satire, propaganda, hoax, or credible news, Asr et al., 2018 [3], a large comprehensive dataset of fact-checked news articles taken from Snopes and Politifact, and the BuzzFeedUSE dataset [8], a collection of veracity labels for Facebook links. These datasets have been the foundation for many ML models to detect fake news. Many different ML models have been created such as the deep diffusive network model (Zhang et al., 2018) [21] or the Naïve Bayes model (Granik et al., 2017) [6]. However, this existing research used unbalanced data sets (Granik et al., 2017), small data sets (Perez-Rosas et al., 2017) [12], or data sets from a single domain (Zhang et al., 2017). Unbalanced datasets between fake and real news may contain significant differences in subject matter, the number of articles, or the date of publishing, contributing to skewed results. Despite the many ML models available, no algorithm fully controlled their lurking variables nor determined the features of fake news. This paper introduces a robust classification model which aids in distinguishing the characteristics that are essential in the composition of fake news. 

\section{Methodology}
\subsection{Data Collection} Due to the data-dependent nature of ML classifiers, collecting unbiased data is essential to its success.  To eliminate all confounding variables, the datasets of fake and credible articles need to be similar in subject matter, writing style, size, publication date (2 weeks), and political lean. This allows the classifier to classify based on word distribution and other linguistic features of the articles. Data used in preparation was obtained from 15 credible [10] and 15 fake news sources [2]. Each credible article was meticulously hand labelled by fact-checking and cross-referencing other verifiable sources. These sources were labelled as credible by established fact-checkers such as PolitiFact or Snopes. Fake news sources were taken from The Fake News Codex [2], a collection of websites known to publish fake news. 

\subsection{Text Scraping and Cleaning}
Text scraping was done through a Java library called JSoup which scrapes text based on its HTML tag. The contents of the articles were scraped from the HTML paragraph element \textless{}p\textgreater{}. However, many extraneous words in the paragraph element were also scraped such as the report’s location, or the journalist’s name. These statements were removed from classification. For the classifier to function at its optimal level, each scraped word from the websites needed to be in their inflectional forms because words with the same root (decided, decide, deciding) should be treated as the same word (decide). [17] To achieve this, the text was stripped of HTTPS, removed of non-alphabetical characters, converted into lowercase, and its words stemmed to their root word. [13] 
\subsection{Data Preprocessing} 
This classifier is to be trained based on the word distributions of each article. However, it is only important that high-frequency words appear in the article. For example, if a word appeared once or twice, it should not hold any weight in classification.  
Since chi-square tests statistical analyses were conducted on the optimal data set, words that appeared at least 6 times in List A were chosen to be part of the list (List B) of common words. [1] 
\newline \newline 
Words are divided into two categories: function words, which signify grammatical relationships (the, to, a...), and content words, which have meaning (green, Science...).  A list of function words found on Semantic Similarity [16] was removed from List B.  After removal, List B contained 1000 common words.
\newline \newline 
To generate article vectors for classification, each article was assigned 1000 parameters, ranked in terms of importance to indicate its keywords. The order of the articles was randomized to ensure fair representation of all domains. To classify fake and credible news, articles are compared to each other based on these parameters.  

\subsection{Classification}
In order to balance domains, the data were randomly split into training, validation and testing categories. 60\% of the data was used for training, 20\% for validation, and the remaining 20\% for testing. 21 classification models were trained and tested at 95\% variance. The top 3 models (based on their classification performance on the validation set) per word count were tested with the testing set to find the overall best model. Furthermore, the errors made by the classifier were analyzed in order to eliminate mistakes from future classifiers and improve upon their accuracy. 

\subsection{Feature Analysis} Recognizing the most important features for classification demonstrates the differentiation between fake news and credible news. In essence, the features of an article are dependent on its words, sentences, and paragraphs. A word’s features include length and a part of speech.  A sentence’s or paragraph’s features are its length, its tone (sentiment) and its degree of formality and authenticity. Lastly, the overall word distribution in these articles was also very important. The word’s part of speech and the overall sentiment (tone) of the text were extracted with the Stanford CoreNLP library [17] [19]. The degrees of formality and authenticity were extracted using the Linguistic Inquiry and Word Count (LIWC) and Receptivi API [11]. These features were analyzed, and various statistical tests were performed to find their significance and weight (Table 2, 3, 4, 5, 6).

\section{Results}

\begin{table}[htb]
\centering
\caption{Classification Model by Number of Words}
\label{t:observed_psrs}
\begin{tabular}{rcccc}
\noalign{\smallskip} \hline \hline \noalign{\smallskip}
\#of Words & Model & Recall (\%) & F1 Score (\%) & Classification Accuracy (\%)\\
\hline
30              & Quadratic SVM & 52     & 57       & 54                      \\
60              & Bagged Trees  & 75     & 75       & 77                      \\
90              & Linear SVM    & 82     & 82       & 82                      \\
120             & Linear SVM    & 83     & 82       & 83                      \\
90 EA           & Linear SVM    & 82     & 82       & 87                      \\
120 EA          & Linear SVM    & 83     & 82       & 87
\\ 
\hline
\end{tabular}
\end{table}
\begin{table}[htb]
\centering
\caption{Part of Speech Distribution in Real and Fake News}
\label{t:observed_psrs}
\begin{tabular}{rcccc}
\noalign{\smallskip} \hline \hline \noalign{\smallskip}
Part of Speech & Z- Score & P-value            \\
\hline
Adjective      & 2.561    & 0.0104             \\
Adverb         & 7.225    & \textless{}0.00001 \\
Noun           & -5.428   & \textless{}0.00001 \\
Pronoun        & 1.607    & 0.108              \\
Verb           & 0.989    & 0.336             
\\ 
\hline
\end{tabular}
\end{table}
\begin{table}[htb]
\centering
\caption{Sentiment Score Distribution in Real and Fake News}
\label{t:observed_psrs}
\begin{tabular}{rcccc}
\noalign{\smallskip} \hline \hline \noalign{\smallskip}
Sentiment & Z- Score & P-value            \\
\hline
Very Positive & 0.886    & 0.375   \\
Positive      & 0.0307   & 0.976   \\
Neutral       & 1.154    & 0.248   \\
Negative      & -1.086   & 0.278   \\
Very Negative & -0.992   & 0.321          
\\ 
\hline
\\ 
\end{tabular}

\centering
\caption{Convention Lengths Distributions in Real and Fake News}
\label{t:observed_psrs}
\begin{tabular}{rcccc}
\noalign{\smallskip} \hline \hline \noalign{\smallskip}
Length & Z- Score & P-value            \\
\hline
Word          & -0.250   & 0.802   \\
Sentence      & 0.651    & 0.515   \\  
\hline
\end{tabular}
\end{table}
\begin{table}[htb]
\newcommand{\ra}[1]{\renewcommand{\arraystretch}{#1}}
\centering
\caption{LIWC Summary Variable Distribution in Real and Fake News}
\ra{1.05}
\begin{tabular}{@{}ccccccc@{}}
\toprule 
& \multicolumn{2}{c}{Tone Authenticity} & \phantom{a}& \multicolumn{3}{c}{Emotional Tone} \\
\cmidrule{2-7}
&Z-score & p-value && Chi-square ($\chi^2$) & df & p-value \\ \midrule
\phantom{c} & 1.863 & 0.0625 && 630.5 & 576 & 0.0574\\
\phantom{c} & -1.122& 0.262 && 398.3 & 380 & 0.249\\
\bottomrule
\end{tabular}
\end{table}
\begin{table}
\centering
\caption{Word Distributions in Real and Fake News}
\label{t:observed_psrs}
\begin{tabular}{rcccc}
\noalign{\smallskip} \hline \hline \noalign{\smallskip}
Chi-Squared Test & Two Sample Z-Test   & Most influential Words\\
\hline
p-value \textless 0.0001  & p-value = 0.0437 & Clinton, gun Obama, city \\ \hline
\\
\end{tabular}
\centering
\caption{Minimum Frequency of Words in List A per Word Count}
\label{t:observed_psrs}
\begin{tabular}{rcccc}
\noalign{\smallskip} \hline \hline \noalign{\smallskip}
\# of Words & Minimum Frequency of Words \\
\hline
30          & 4                          \\
60          & 5                          \\
90          & 6                          \\
120         & 6 \\  \hline    
\end{tabular}
\end{table}

\section{Discussion}
\subsection{Model Performances} 
The Linear SVM model initially reached a testing accuracy of 83\%. The categories of Type I and Type II errors made by the classifiers were analyzed and split into 3 groups (Figure 1, Figure 2). In the testing set, 200 credible articles were checked to ensure that each article was relevant to the series of topics while simultaneously replacing irrelevant ones with a different article from the same domain and time period. 200 fake articles were also re-fact checked to ensure that the contents were false. In total, 13 fake articles and 21 credible articles were replaced. After this error analysis, both classifying accuracies on the testing set for 90 and 120 words converged at 87\%. (Table 1.) 
\begin{center}
    \includegraphics[scale = 0.45]{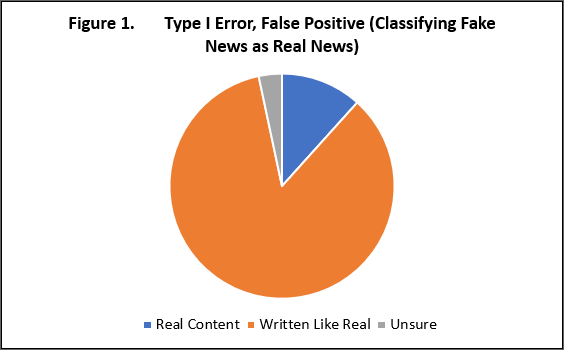}
    \includegraphics[scale = 0.45]{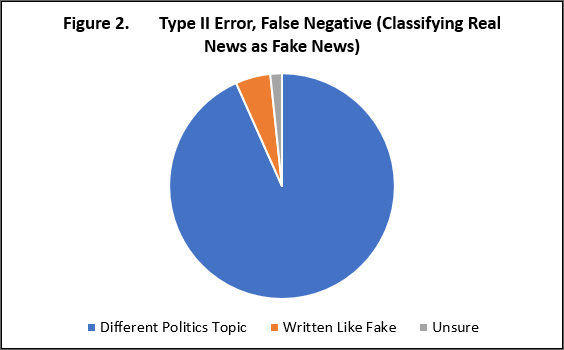}
\end{center}
This model (90 words) displayed a greater accuracy than both Granik et al. and Zhang et al. even though both utilized skewed data sets. The model classified fake news with an accuracy of 88\% and credible news at 85\%. Overall the classifiers obtained a recall rate of 82\%. This is important as mislabeling credible data could cause the user to trust a deceptive website, increasing their intake of fake news.
\subsection{Feature Analysis} 
Based on this dataset, adverbs were used 40\% more in fake news articles (p$<0.00001$) probably to give emphasis to deceptive information. The higher usage of nouns in credible news could be attributed to credible news presenting more objective information since nouns tend to not hold any emotion. (Table 2.) 
\\ \\ The p-values of sentiment analysis were not statistically significant. This data was supported by the LIWC summary variable of emotional tone which displayed no significant statistical difference between credible and fake news. This contradicts the notion that fake news is emotionally colored. (Table 3.) The differences in word length and sentence length were also statistically insignificant disproving the notion that credible news tends to use more complex language than fake news. (Table 4.) However, LIWC data showed that the authentic tone of fake news was significantly higher than that of credible news. (Table 5.) This deviates from the stereotypical notions of fake and credible news and can be attributed to the fact that fake news puts in more effort to make their content seem authentic. The statistical differences in word distribution were also very significant (Table 6, Figure 3, Figure 4). The model appears to use all of these distinctions in features to classify fake and real news.
\begin{center}
\includegraphics[scale = 0.4]{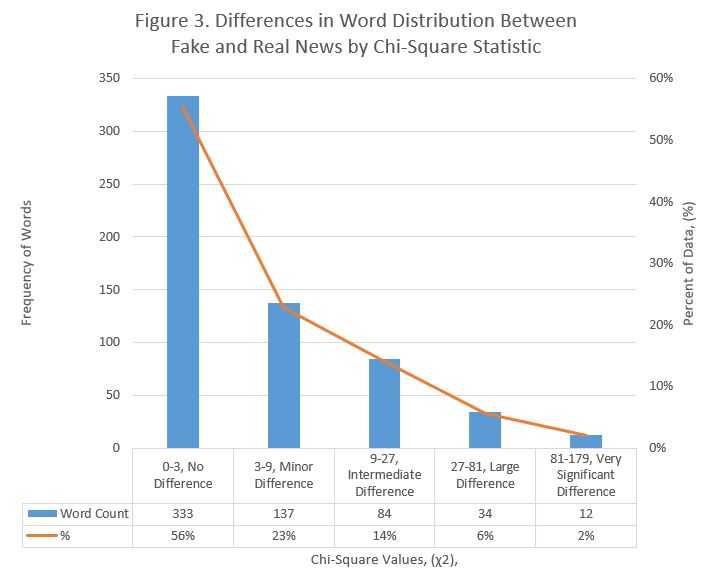}
    \includegraphics[scale = 0.4]{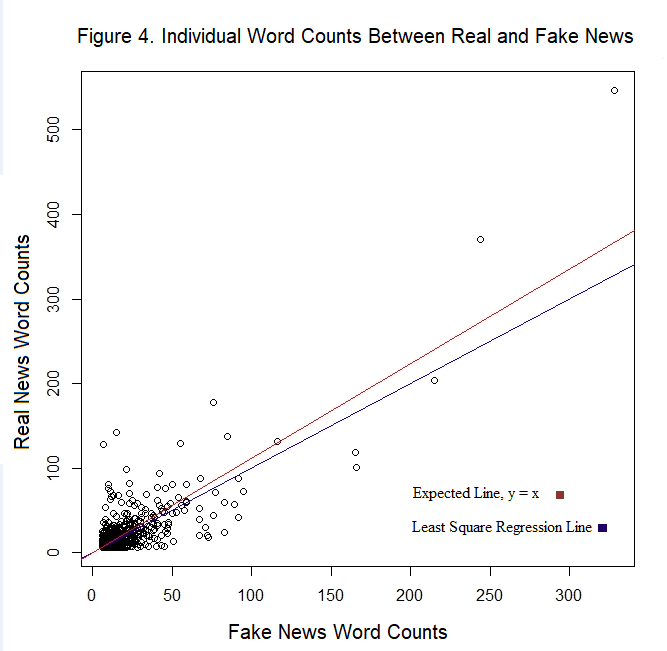}
\end{center}
\subsection{Topic Distribution}
The topics of the dataset were further analyzed and it was noticed that there were visibly more topics in fake news than real news. This could be the result of the data collection method: Much of the real news was collected by compiling similar news topics that were fact-checked by fact-checkers such as \textit{Snopes} and \textit{Politifact} and labelled as \textit{true}. Thus, there are more clusters of topics and less individual topics. 
\\ \\ 
There was a general trend that if the topics did not overlap between real and fake news, the classifier performed better. This may suggest that the classifier is classifying on topics rather than content. 
\\ 
After close comparisons between fake and real news of the same topic, it is evident that the content in fake news is not the polar opposite of credible news. Instead, they seem to stretch the truth to one end of the political spectrum or adds additional deceptive information. 
\newline \newline 
Overall there were more topics in fake news and so the articles were more spread out and as a result, there were less fake news articles that overlapped with the real news articles. In some topics such as \textit{Mueller investigation} the classification was lackluster while in other topics such as the \textit{Parkland shooting}, the performance was very strong. This difference can arise due to the divisiveness of a certain topic. There appears to be a general trend between the magnitude of the topic and the accuracy of classification. Topics that are of greater magnitude in terms of relevance tend to have a greater exposure to the general public, with greater exposures come more diverse perspectives. 

\section{Conclusion} 
This paper presents a high accuracy ML classifier to classify the validity of online news articles to be Credible or Fake based on word distributions and other linguistic and stylistic features. The best performing model by overall classification accuracy on the testing set was Linear SVM reaching 87.0\%. This model performed better than existing classifiers even with unbalanced datasets such as Granik et al. [6] and Zhang et al. [20]. The optimal number of words (X) in an article to test was found to be 90 words since the accuracy converged after that point. Additionally, by analyzing the model and its features, this project provides insight into the specific features of fake news. This classifier appears to use the differences in word distribution, levels of tone authenticity and the frequency of different parts of speech. In particular, the frequency of adverbs, adjectives, and nouns showed very significant statistical differences between real and fake news.

\section{Future Work}
This model presented also provides a few limitations which reduce the inferences that can be drawn. This classifier requires large amounts of data to stay updated to each news cycle and produce optimal results. Therefore, finding the minimum amount of data required to still attain high accuracy rates could drastically reduce the time to prepare data for the classifier. An automated data collection system can be implemented to quickly and cost-effectively collect data. Moreover, this classifier is binary, while news can be partially credible or fake. Thus, adding extra dimensions may present further useful information. This model may be coded into a chrome extension and aid the general public in making better-informed decisions. Further applying a successful classifying filter to news websites or social media outlets can help reduce the large amounts of misinformation circulating the Internet.
\section{Acknowledgements} 
I thank Dr. Maite Tobaoda, a linguistics professor at Simon Fraser University. She has given me the opportunity to take part in a discourse processing lab. I thank Dr. Fatemeh Torabi Asr, a post-doctorate at Simon Fraser University who has kindly mentored me on data collection and project development and given me invaluable advice on extensions to my project. I thank
Dr. Joan Hu, a statistics professor at Simon Fraser University who has advised me on statistical procedures. I thank William Chow for his help in editing this paper.
\setlength{\parindent}{4em}
\setlength{\parskip}{1em}
\renewcommand{\baselinestretch}{1.0}


\begin{thebibliography}{}
\bibitem{1}The Chi-square test of independence. (n.d.). Retrieved from \url{https://www.ncbi.nlm.nih.gov/pmc/articles/PMC3900058/}
\bibitem{2}The Fake News Codex. (n.d.). Retrieved from \url{http://www.fakenewscodex.com/}
\bibitem{4}Fatemeh Torabi Asr and Maite Taboada (2019) MisInfoText. A collection of news articles, with false and true labels. Dataset.
\bibitem{4}Function word lists. (2013, November 17). Retrieved from \url{http://semanticsimilarity.wordpress.com/function-word-lists/ }
\bibitem{5}Gilda, S. (2017). Evaluating machine learning algorithms for fake news detection. 2017 IEEE 15th Student Conference on Research and Development (SCOReD). doi:10.1109/scored.2017.8305411
\bibitem{6}Granik, M., \& Mesyura, V. (2017). Fake news detection using naive Bayes classifier. 2017 IEEE First Ukraine Conference on Electrical and Computer Engineering (UKRCON). doi:10.1109/ukrcon.2017.8100379
\bibitem{7}How Fake News Goes Viral: A Case Study. (2017, December 22). Retrieved from \url{https://www.nytimes.com/2016/11/20/business/media/how-fake-news-spreads.html}
\bibitem{8}Hyperpartisan Facebook Pages Are Publishing False and Misleading Information at An Alarming Rate. (2016, October 20). Retrieved from \url{https://www.buzzfeednews.com/article/craigsilverman/partisan-fb-pages-analysis} 
\bibitem{9}In Washington Pizzeria Attack, Fake News Brought Real Guns. (2018, January 20). Retrieved from \url{https://www.nytimes.com/2016/12/05/business/media/comet-ping-pong-pizza-shooting-fake-news-consequences.html}
\bibitem{10}List of Websites Used for Real News. (2019, March 1). Retrieved from \url{https://drive.google.com/file/d/1bbi4t0MX31TTzVNWibsP0qtsqPd1qHyC/view?usp=sharing} 
\bibitem{12}Pennbaker, J. W., Boyd, R. L., Jordan, K., \& Blackburn, K. (2015). The Development and Psychometric Properties of LIWC2015. Retrieved from \url{https://repositories.lib.utexas.edu/bitstream/handle/2152/31333/LIWC2015_LanguageManual.pdf}
\bibitem{13}Perez-Rosas, V., Kleinberg, B., Lefevre, A., \& Mihalcea, R. (n.d.). Automatic Detection of Fake News. Retrieved from \url{https://arxiv.org/abs/1708.07104}
\bibitem{14}Porter Stemmer Online. (2012, September 7). Retrieved from \url{http://9ol.es/porter_js_demo.html} 
\bibitem{15}Precision and recall. (2018, November 3). Retrieved June 3, 2018, from \url{http://en.wikipedia.org/wiki/Precision_and_recall }
\bibitem{16}Rashkin, H., Choi, E., Jang, J. Y., Volkova, S., \& Choi, Y. (2017). Truth of Varying Shades: Analyzing Language in Fake News and Political Fact-Checking. Proceedings of the 2017 Conference on Empirical Methods in Natural Language Processing. doi:10.18653/v1/d17-1317
\bibitem{17}Semantic Similarity Function word lists. (2013, November 17). Retrieved from \url{https://semanticsimilarity.wordpress.com/function-word-lists/}
\bibitem{18}Socher, R., Perelygin, A., Wu, J. Y., Chuang, J., Manning, C. D., Ng, A. Y., \& Potts, C. (2013). Recursive Deep Models for Semantic Compositionality Over a Sentiment Treebank. Retrieved from \url{https://nlp.stanford.edu/~socherr/EMNLP2013_RNTN.pdf}
\bibitem{19}Stemming and lemmatization. (2009, April 7). Retrieved from 
\url{http://nlp.stanford.edu/IR-book/html/htmledition/stemming-and-lemmatization-1.html}
\bibitem{20}Toutanova, K., Klein, D., Manning, C. D., \& Singer, Y. (2003). Feature-rich part-of-speech tagging with a cyclic dependency network. Proceedings of the 2003 Conference of the North American Chapter of the Association for Computational Linguistics on Human Language Technology - NAACL '03. doi:10.3115/1073445.1073478
\bibitem{21}Wang, W. Y. (2017). "Liar, Liar Pants on Fire": A New Benchmark Dataset for Fake News Detection. Proceedings of the 55th Annual Meeting of the Association for Computational Linguistics (Volume 2: Short Papers). doi:10.18653/v1/p17-2067 
\bibitem{22} Zhang, J., Cui, L., Fu, Y., \& Gouza, F. B. (n.d.). Fake News Detection with Deep Diffusive Network Model. arXiv.org. Retrieved from \url{https://arxiv.org/abs/1805.08751 }


\end{thebibliography}
\end{document}